\documentclass[11pt,a4paper]{article}
\usepackage[hyperref]{acl2020}
\usepackage{times}
\usepackage{latexsym}

\usepackage{graphicx}
\usepackage{multirow}
\usepackage{bm}
\usepackage{amsmath}
\usepackage{amsfonts}
\usepackage{tablefootnote}
\usepackage[switch]{lineno}

\usepackage{microtype}

\aclfinalcopy 

\title{Bridging the Modality Gap for Speech-to-Text Translation}

\author{Yuchen Liu$^{1,2}$, 
        Junnan Zhu$^{1,2}$, 
        Jiajun Zhang$^{1,2}$, 
        \and Chengqing Zong$^{1,2,3}$ \\
        $^{1}$ National Laboratory of Pattern Recognition, Institute of Automation, CAS \\ 
        $^{2}$School of Artificial Intelligence, University of Chinese Academy of Sciences \\
        $^{3}$CAS Center for Excellence in Brain Science and Intelligence Technology, Shanghai, China \\
{\tt \{yuchen.liu, junnan.zhu, jjzhang, cqzong\}@nlpr.ia.ac.cn}}

\date{}

\begin{document}
\maketitle
\begin{abstract}
End-to-end speech translation aims to translate speech in one language into text in another language via an end-to-end way.
Most existing methods employ an encoder-decoder structure with a single encoder to learn acoustic representation and semantic information simultaneously, which ignores the speech-and-text modality differences and makes the encoder overloaded, leading to great difficulty in learning such a model. 
To address these issues, we propose a Speech-to-Text Adaptation for Speech Translation (STAST) model which aims to improve the end-to-end model performance by bridging the modality gap between speech and text. 
Specifically, we decouple the speech translation encoder into three parts and introduce a shrink mechanism to match the length of speech representation with that of the corresponding text transcription. To obtain better semantic representation, we completely integrate a text-based translation model into the STAST so that two tasks can be trained in the same latent space. Furthermore, we introduce a cross-modal adaptation method to close the distance between speech and text representation. 
Experimental results on English-French and English-German speech translation corpora have shown that our model significantly outperforms strong baselines, and achieves the new state-of-the-art performance.
\end{abstract}

\section{Introduction}
\label{intro}
Speech-to-Text translation (ST) aims at translating speech in one language into text in another language, which can be widely applied to conference speech, cross-border service, international business talk, academic forum, etc. Most existing approaches to speech-to-text translation are based on the pipeline paradigm, which first transforms the speech into text via an automatic speech recognition (ASR) system and then translates the transcribed text into the target language by a text-based machine translation model \cite{ney1999speech,kanthak2005novel,mathias2006statistical}. 

Recently, the end-to-end speech-to-text translation model has attracted more attention due to its advantages over the pipeline paradigm, such as low latency, alleviation of error propagation, and fewer parameters \cite{weiss2017sequence,berard2018end,bansal2018pre,jia2019leveraging,sperber2019attention}. Previous studies employ an encoder-decoder structure to directly learn the mapping relationship between the speech input and text sequences in the target language. However, modality differences between speech and text result in the difficulty of training such a model, making its performance usually much inferior to the corresponding text-based neural machine translation (NMT) model.

We observe that there are three main modality differences between speech and text which affect the speech representational capacity. First, the length of frame-level speech features is much longer than that of the corresponding text data, which hinders the model from learning alignments between the input and the output sequence. Second, text data are generally digitalized into trainable embeddings, while speech features are calculated by hand-crafted filters and fixed during training, resulting in a lack of semantic information. Third, text data contain less noise with low uncertainty, while the speech is variational and easily affected by factors like the speech rate of the different speaker, silence, and noise, which leads to the poor robustness of the model. To make matter worse, the encoder in the ST model is overloaded 
since it requires to both learn acoustic information and extract semantic knowledge from the speech input simultaneously.

In order to release the burden of speech translation encoder and help it learn better representation, the text representation learned by the NMT model can be utilized as a guidance for speech representation learning. To achieve that, we propose a Speech-to-Text Adaptation for Speech Translation (STAST) model which aims to improve the end-to-end model performance by bridging the modality gap between speech and text. Specifically, the representations of two modalities need to be consistent in terms of length and share in the same latent space. 
Therefore, we decouple speech translation encoder into three parts, including an acoustic encoder, a shrink mechanism, and a semantic encoder. The shrink mechanism can filter redundant hidden states based on the spike-like label posterior probabilities generated by the CTC module, which has the ability to solve length inconsistency problems. 
To ensure the representation of two modalities being in the same semantic latent space, we apply the multi-task learning method and completely integrate the MT model into the ST model by sharing the semantic encoder and decoder. 
To further close the distance of two representations, we apply a cross-modal adaptation method, which can afford speech representation more semantic information. 
Experimental results on two speech translation corpora have illustrated the effectiveness of our proposed model. 

The contributions of this paper are as follows:
\begin{itemize}
\item We propose the STAST model which can improve speech translation performance by bridging the modality gap between speech and text.
\item Experimental results have demonstrated that our proposed method outperforms existing methods and achieves the new state-of-the-art performance.
\item Our method can easily leverage extra data and has shown especially superiority in low-resource scenarios.
\end{itemize}

\section{Related Work}
\textbf{Speech Translation.} Traditional studies on speech translation are based on the pipeline paradigm which consists of an ASR model and an MT model \cite{ney1999speech,kanthak2005novel,mathias2006statistical}. Focusing on how to improve the translation robustness on spoken language domain and combine the separate ASR and MT models, previous studies propose lattice-to-sequence models, synthetic data augment, and domain adaptation techniques \cite{lavie1996multi,waibel2008spoken,sperber2017neural,sperber2019self}. 

Recently, studies based on the end-to-end paradigm have emerged rapidly due to its advantages, such as lower latency, alleviation of error propagation, and fewer parameters. 
\citet{1999The} presume that it is possible to implement such end-to-end speech translation with the development of memory, computation speed, and representation methods. Then \citet{berard2016listen} give the first proof of the potential for an end-to-end ST model. Since then, pre-training, multitask learning, attention-passing, and knowledge distillation have been applied to improve the end-to-end model performance \cite{anastasopoulos2016unsupervised,duong2016attentional,weiss2017sequence,berard2018end,sperber2019attention,liu2019end,jia2019leveraging,liu2020synchronous}. 

Considering the difficulty of modeling this cross-modal cross-lingual task in a single model, recent studies propose novel model structures or auxiliary tasks to enhance the model capability. \citet{wang2020bridging} propose TCEN model which connects speech encoder and NMT encoder in tandem. This model aims to bridge the gap between pre-training and fine-tuning by reusing every sub-net. Considering the length inconsistency between speech encoder outputs and word embeddings, they lengthen the source text sentence by adding word repetitions and blank tokens to mimic the CTC output sequence. However, this process needs to train an extra sequence-to-sequence model and introduces much noise to the NMT model. They later propose a curriculum pre-training method which integrates two elementary courses to enable the encoder to understand the meaning of a sentence and map words in different languages \cite{wang2020curriculum}. However, it conducts the force-alignment between the speech and the source word as well as the source-to-target word alignment which needs to train an extra ASR model and may introduce alignment errors. In total, these studies do not well solve the modality gap between speech and text. Consequently, much valuable semantic information learned by text-based MT model cannot be applied to the ST model, which limits the latter performance.

\noindent \textbf{Cross-modal Adaptation.} The paradigm of minimizing the difference between two models can be adopted in the transfer learning, where the knowledge embedded in one model can be transferred to another model, including output probabilities \cite{hinton2015distilling,freitag2017ensemble}, hidden representations \cite{yim2017gift,romero2014fitnets}, and generated sequences \cite{kim2016sequence}. 
Such scheme has been  applied in a variety of tasks, such as image classification \cite{hinton2015distilling,li2017learning,yang2018knowledge,anil2018large}, speech recognition \cite{hinton2015distilling} and natural language processing \cite{freitag2017ensemble,kim2016sequence,tan2019multilingual}.
Some related studies have attempted to transfer knowledge from the text model to the speech model by cross-modal adaptation. 
\citet{cho2020speech} and \citet{denisov2020pretrained} adopts this method on spoken language understanding task, while \citet{liu2019end} apply it to speech translation task. However, they only transfer the knowledge from the output of the last model layer, ignoring the representation gap between speech and text modality.

\section{Method}
In this section, we first introduce the architecture of our proposed STAST model. Focusing on bridging the representation gap between speech and text modality, it decouples the encoder into three parts to transcribe the speech and extract semantic representation separately.
To further close the semantic gap between two modalities, we apply a cross-modal adaptation method. Finally, we will give the training strategy for this task.

\subsection{Problem Formulation}
The speech translation corpus usually includes triplets of speech, transcription, and translation, denotes as $\mathcal{D}_{ST}=\{\bm{(s,x,y)}\}$, where $\bm{s}=[s_1,\cdots,s_{T_s}]$ is a sequence of speech features which are converted from the speech signals, $\bm{x}=[x_1,\cdots,x_{T_x}]$ is the corresponding transcription in source language, and $\bm{y}=[y_1,\cdots,y_{T_y}]$ denotes the translation in target language. $T_s$, $T_x$, $T_y$ are the length of speech features, transcription and translation, respectively, where $T_s \gg T_x$.
An extra ASR dataset $\mathcal{D}_{ASR}=\{\bm{(s',x')}\}$ can be leveraged for pre-training the ST model.

\begin{figure*}[t]
\centering
\includegraphics[width=2.0\columnwidth]{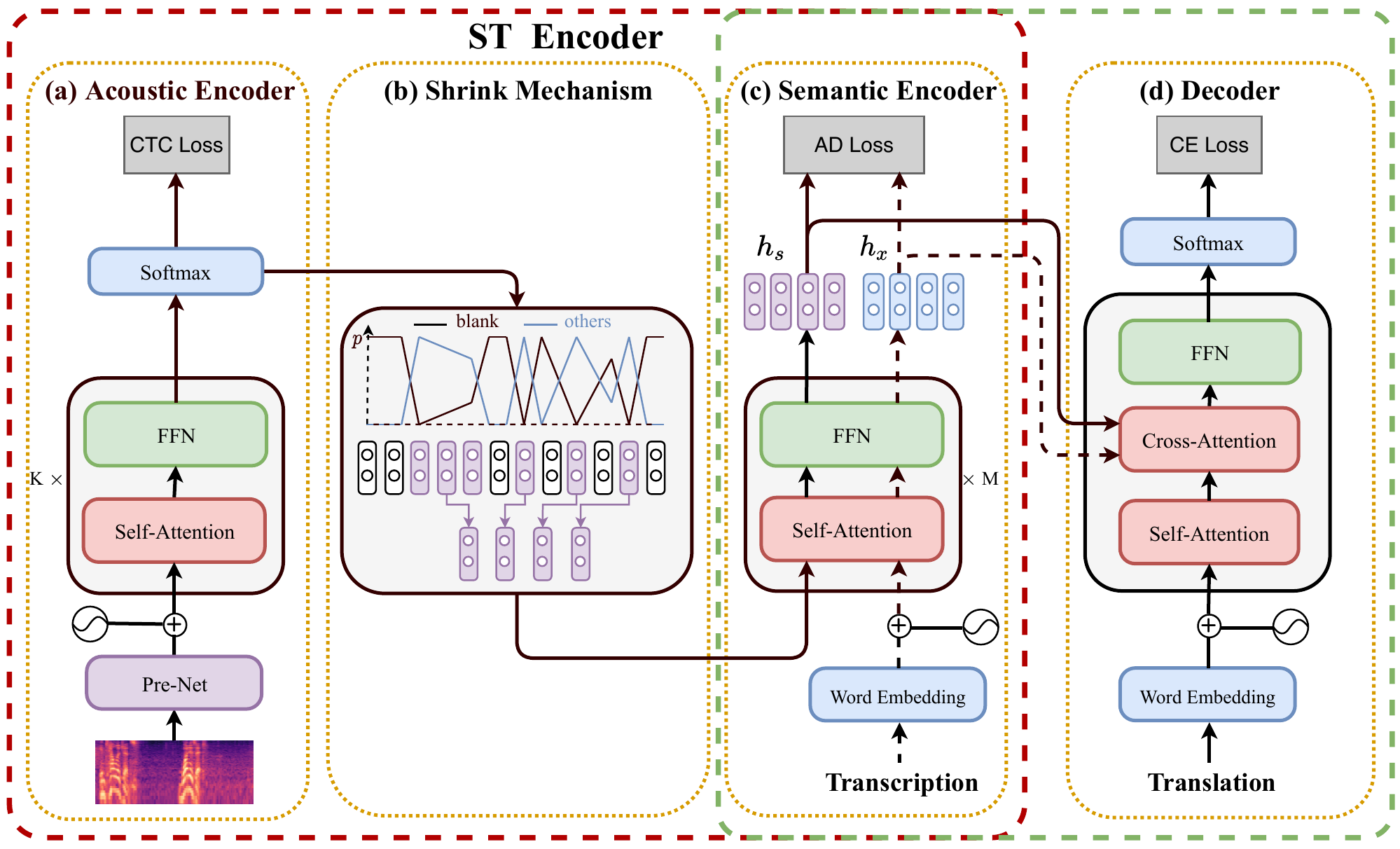}
\caption{Overview of our proposed STAST model. The speech translation encoder is decoupled into three parts, including \textbf{(a)} acoustic encoder, \textbf{(b)} shrink mechanism, and \textbf{(c)} semantic encoder, where \textbf{(c)} semantic encoder and \textbf{(d)} decoder form the integrated NMT model. Black dotted lines denote that process is only used during training. For the sake of clarity, residual connection and layer normalization are not shown.}
\label{fig:model}
\end{figure*}

\subsection{Model Architecture}
STAST model adopts the encoder-decoder framework, as shown in Figure \ref{fig:model}. Both the encoder and decoder adopt Transformer \cite{vaswani2017attention} as the basic model structure since its superior property.
Compared with the encoder in the ASR model which only needs to learn acoustic knowledge, the encoder in the end-to-end speech translation models is overloaded which requires to learn both acoustic and semantic knowledge of the source speech.
To release its burden, we decouple the ST encoder into three parts, i.e. an acoustic encoder concatenated by a shrink mechanism, and a semantic encoder.
Specifically, the acoustic encoder adopts a CTC module to learn speech representation and to predict the source transcription, which plays the role of an ASR model. To ensure the length of speech representation and text representations being consistent, a shrink mechanism is applied on the output of the acoustic encoder to filter redundant states based on the spike-like label posterior probabilities which are generated by the CTC module. This process can significantly reduce the length of speech input, as well as reserve most of the meaningful information in the source speech. Then the semantic encoder encodes the hidden states corresponding to the non-redundant positions to obtain better semantic representation, based on which decoder with an attention module generates the final translation. We introduce each part in detail as follows.

\noindent\textbf{Acoustic Encoder.} Since we decouple the ST encoder to executive different functions, the acoustic encoder here is mainly used to learn acoustic knowledge. 
It takes as input the sequence of speech features $\bm s$. 
For speech inputs, we first employ a speech pre-net to extract speech features, which is a linear layer here. The feature dimension is converted into the model hidden size $d_{\rm model}$. Then the acoustic representation is extracted by multiple stacked self-attention layers. The above process can be formalized as :
\begin{align}
\tilde{\bm{s}} &= {\textrm{Pre-Net}}(\bm{s}) \\
\bm{h} &= {\textrm{Acoustic\_Encoder}}(\bm{\tilde{s}}) \nonumber \\
&= {\textrm{FFN}}({\textrm{Self\_Attention}}(\bm{\tilde{s}}))
\end{align}
\textbf{CTC Module.} To predict source transcriptions, we adopt the Connectionist Temporal Classification (CTC) \cite{graves2006connectionist} module on the output of the acoustic encoder.
Given the hidden states $\bm{h}$ generated by the acoustic encoder, a softmax classification layer is applied to predict a CTC path $\bm{\pi}=[\pi_1,\pi_2,\cdots,\pi_{T_s}]$,  where $\pi_t \in \mathbb{V}\cup$``-" denotes the label predicted by softmax layer at each step $t$, $\mathbb{V}$ is the vocabulary and ``-"  is a blank label. Then the distribution over a CTC path $\bm \pi$ can be calculated as the probability of a sequence of conditional independent outputs: 

\begin{align}
p(\bm \pi|s) = \prod_{t=1}^{\textrm{T}_s}p(\bm {\pi}_t|\bm {s})
=\prod_{t=1}^{\textrm{T}_s} \textrm{softmax}(\bm{ W}_{\textrm{ctc}}^T \times h_t)
\end{align}

where $\bm{ W}_{\textrm{ctc}}\in \mathbb{R}^{d\times (|\mathbb{V}|+1)}$ is a trainable matrix in the classification layer.

Note that the CTC path $\bm{\pi}$ is a many-to-one mapping of the source transcription $\bm{x}$ by allowing occurrences of blank or consecutively repeated label.
For example, let $\mathbb{B}$ denotes the mapping from CTC paths to the transcription sequence, then $\mathbb{B}(aa-ab-)=B(a-abb-)=aab$, where ${\bm \pi}_1=aa-ab-$ and $\bm {\pi}_2=a-abb-$. 
There exists many legal CTC paths for each transcription $\bm x$. 
Therefore, the conditional probability of each transcription $\bm x$ can be modeled by summing all the paths corresponding to it.

\begin{align}
p_{ctc}(\bm{x|s}) = \sum_{\bm{\pi} \in \mathbb{B}^{-1}(\bm{x})}p(\bm{\pi|s})  
\end{align}

\noindent where $\mathbb{B}^{-1}(\bm x)$ denotes the set of all legal CTC paths corresponding to a transcription $\bm x$.
Finally, the objective training function of CTC loss is defined as,

\begin{align}
\mathcal{L}_{\rm CTC} = -\sum_{(\bm {s,x})\in \mathcal{D}} \log p_{\rm ctc}(\bm {x|s})
\end{align}


\noindent\textbf{Shrink Mechanism.} Note that the length of CTC output is the same as the input speech feature, which is still much larger than that of the corresponding source text. To bridge the length gap between speech and text representation, we apply a shrink mechanism. 

As mentioned above, CTC paths are variation of the source transcription by allowing occurrences of blank tokens and repetitions. However, blank and repeated tokens do not contain any useful information and hinder the linguistics modeling. We assume that the triggered encode state sequence contains prior information of original speech input. Therefore, we only extract the encoded states $\bm{h}$ in the acoustic encoder which corresponds to the CTC spike, as shown in the middle part of Figure \ref{fig:model}. Specifically, we treat the label which has the largest value after the softmax layer in each state $h_t$ as the predicted label. Then, we only extract the hidden state whose label does not correspond to the blank label or consecutively repeated label and mask other states, which is similar with~\citeauthor{yi2019ectc}~\shortcite{yi2019ectc} and~\citeauthor{tian2020spike}~\shortcite{tian2020spike}. 
Note that this process does not affect the back propagation of gradient. Then the shrunk hidden state can be formalized as follows, where $\rm Idx$ denotes the corresponding index of token in the vocabulary.

\begin{small}
\begin{align}
\tilde{\bm{h}} = [h_i \in \bm{h} | \mathop{\arg\max}_{\rm Idx}\ {\rm softmax}(\bm{ W}_{\textrm{ctc}}^T \times h_i) \neq {\rm Idx}(``-")]
\end{align}
\end{small}

\noindent\textbf{Semantic Encoder and Decoder.} The shrunk hidden state contains acoustic information but still lacks semantic knowledge.
To obtain better semantic representation, we apply semantic encoder to encode the shrunk hidden states by another multiple stacked self-attention layers.

\begin{align}
\bm{h_s} = {\textrm{FFN}}({\textrm{Self\_Attention}}(\bm{\tilde{h}}))
\end{align}

\noindent The decoder also follow the basic network structure of Transformer, it first adopts self-attention layers on target embeddings and then attends to the output of semantic encoder by cross-attention layers. Followed by feed-forward layer, the target token is predicted through a softmax layer based on the output of the decoder $\bm{h_d}$. The above process can be formalized as follows:

\begin{align}
\bm{h_y} = &\textrm{Embedding}(\bm{y}) \\
\bm{h_d} = &{\textrm{FFN}}({\textrm{Cross\_Attention}}( \\ \nonumber
&{\textrm{Self\_Attention}}(\bm{h_y}), \bm{h_s}))
\end{align}

\noindent Finally, the distribution probability over a sequence of target tokens is calculated.

\begin{align}
p(\bm{y|s}) = {\rm softmax}(\bm{W_t}^T \times \bm{h_d})
\end{align}

\noindent where $\bm{W_t} \in \mathbb{R}^{d\times(|\mathbb{V}|+1)}$ is the trainable weight matrix in the softmax layer which is shared with $\bm{W}_{ctc}$ in CTC module.
Then objective function of ST task can be calculated by the cross-entropy loss as follow.

\begin{align}
\mathcal{L}_{ST}=-\sum_{(\bm{s,y}) \in \mathcal{D}}\log p(\bm{y|s})
\end{align}

\noindent \textbf{Integrated NMT Model.} Since the NMT model has the same structure with the combination of semantic encoder and decoder, we integrate the whole NMT model into STAST. In NMT model, the source transcription $\bm x$ is first embedded into word representation by looking up the embedding weight $\bm W_s$. Then semantic encoder extracts high-level semantic text representation $\bm h_x$, based on which the decoder performs translation task.
To map speech representation and text representation into the same semantic space, we share the parameters of semantic encoder and decoder. Meanwhile, we also share the weight $\bm{W}_{ctc}$ of softmax layer in acoustic encoder with the source word embedding weight and the weight $\bm{W_t}$ of the decoder softmax layer to constrain the space gap, which means $\bm{W}_{ctc}=\bm{W_s}=\bm{W_t}$.
Then the objective function of the MT task can be calculated by the cross-entropy loss as follow.

\begin{align}
\mathcal{L}_{MT}=-\sum_{(\bm{x,y}) \in \mathcal{D}}\log p(\bm{y|x})
\end{align}

\subsection{Cross-Modal Adaption}
\label{sec:distillation}
To further make the representation of speech and text modality closer, we propose a cross-modal adaptation method.
Specifically, the cross-modal adaptation method is applied to transfer the semantic knowledge from text representation to speech representation.
For each utterance, the semantic encoder encodes the shrunk output of acoustic encoder into semantic speech representation $\bm h_s$ and encodes word embeddings into text representation $\bm h_x$, respectively. We take advantages of text representation $\bm h_x$ as a regulation to constrain the space of speech representation $\bm h_s$ during training. The supervision is implemented by minimizing the distance between two representations. We propose two adaptation methods, including sentence-level adaptation and word-level adaptation. The loss function can be calculated as:

\begin{equation}
\mathcal{L}_{\rm AD}=\left\{
\begin{array}{lll}
	\sum_{(\bm{s,x}) \in \mathcal{D}} {\rm MSE}(\overline{h}_s, \overline{h}_x)  &  &  \small \textrm{sequence-level} \\
	\sum_{(\bm{s,x}) \in \mathcal{D}} {\rm MSE}(\bm{h_s, h_x})       &      & \small \textrm{word-level}
\end{array} \right. 
\end{equation}

where MSE is mean-squared error loss function used to evaluate the difference between the representation of speech and text, 
$\overline{h}_s$ and $\overline{h}_x$ are the average of two contextual representations.

\subsection{Training Process}
The final training objective function is the sum of four parts, including the CTC loss $L_{ctc}$, the cross-entropy loss for ST task $L_{ST}$, the cross-entropy loss for MT task $L_{MT}$, and the cross-modal adaptation loss $L_{AD}$:
\begin{align}
\label{equ:14}
\mathcal{L}_{ST} = \alpha \mathcal{L}_{CTC} + \beta \mathcal{L}_{ST} + \gamma \mathcal{L}_{MT} + \eta \mathcal{L}_{AD}
\end{align}
where $\alpha$, $\beta$, $\gamma$, and $\eta$ are hyper-parameters, which denote the weight of each loss.

For training strategy, we first train the acoustic encoder by speech-transcription pairs $(\bm{s, x})$ in the ST corpus $\mathcal{D}_{ST}$. Then we use speech-transcription-translation triplets $(\bm{s, x, y})$ to train the ST model and the NMT model by a multi-task learning framework, where transcriptions are only used during training.
The module in our model is very flexible, where the CTC module and the integrated NMT model can perform auxiliary task to obtain better optimized parameters. Therefore, it can be easily trained by extra data, such as the part of acoustic encoder can be trained by extra ASR corpus to obtain better acoustic representation.

\section{Experiments}
\subsection{Datasets}
We conduct experiments on two public speech translation datasets, including Augmented LibriSpeech English-French Corpus \cite{kocabiyikoglu2018augmenting} and the MuST\_C English-German TED Corpus \cite{gangi2019must}.

\noindent\textbf{Augmented LibriSpeech En-Fr.} This corpus is a subset of the LibriSpeech ASR corpus \cite{Panayotov2015Librispeech}, which is automatically aligned with e-books in French. This corpus is from clean audiobooks, which contains quadruplets, including English audios, manual transcriptions, French translations, and the translations obtained by Google Translate. The total audio contains 236 hours of speech. Following previous works \cite{wang2020bridging}, we only use the 100-hour clean training set and concatenate the aligned references with the provided Google translations, resulting in 90K utterances. We validate on the development set (1,071 utterances) and report the model performance on the test set (2,048 utterances).
For pre-training, we use the total LibriSpeech ASR corpus as extra data, which includes 960 hours of speech. 

\noindent\textbf{MuST\_C En-De.}
The MuST\_C corpus is collected from TED talks\footnote{https://www.ted.com}, which includes the English speech, the corresponding transcription, and the target translations in different languages. We conduct experiments on English-German language direction. The speech in this corpus is recorded from the live presentation which contains more noise. The corpus contains a total of 408-hour speech with 234K translation pairs, which is divided into training set (400 hours with 229,703 utterances), development set, and test set. We report case-sensitive BLEU on the dev set (1,423 utterances) and tst-COMMON set (2,641 utterances). 


\subsection{Experimental Settings}
The speech features are 80-dimensional log-Mel filterbanks extracted with a step size of 10ms and window size of 25ms, which are extended with mean subtraction and variance normalization. We adopt dimensionality reduction to downsample one frame every three frames. 

For text data, we apply lowercase, punctuation normalization, and tokenization by Moses scripts\footnote{https://www.statmt.org/moses/}. Punctuations in English transcriptions are removed. We apply the BPE method \cite{sennrich2015neural} on the combination of source and target text to obtain shared subword vocabulary. The number of merge operations in BPE is set to 8K for both tasks. In order to be comparable with other works, we employ case-insensitive  BLEU computed using $\texttt{multi-bleu.pl}$ script\footnote{\url{https://github.com/moses-smt/mosesdecoder/scripts/generic/multi-bleu.perl}} as the evaluation of the translation task.

We use the base configuration of original Transformer \cite{vaswani2017attention}, where the number of self-attention layers in acoustic encoder, semantic encoder, and decoder is 6 with 512-dimensional hidden sizes\footnote{We compare to the ST model with a 12-layer single encoder, whose BLEU is 17.21 on the test set of Augmented LibriSpeech corpus.}, the filter size in feed-forward layer $d_{ff}=2,048$, the residual dropout and attention dropout are $0.1$. We set $\alpha$, $\beta$, $\gamma$, and $\eta$ in Equation \ref{equ:14} to 1.0, 1.0, 1.0, and 1.0 respectively. 
Samples are batched by approximate sequence length of 10,000-frame features.
The STAST model is trained by Adam optimizer \cite{kingma2014adam} on one GPU. We save checkpoints every 1,000 steps and conduct the model average on the last 5 checkpoints as the final model. For inference, we perform beam search with a beam size of 4. Our code will be released after reviews.

\section{Experimental Results}
\subsection{Main Results}
\textbf{Results on Augmented LibriSpeech.}
Following previous work \cite{wang2020curriculum}, we have two settings in this experiment. In the \textbf{Base setting}, we only use the data in the Augmented LibriSpeech corpus. Regarding the \textbf{Expended setting}, we use the 960-hour LibriSpeech ASR corpus as extra data to pre-train the acoustic encoder. Table \ref{tbl:1} presents the results of MT models, pipeline systems, and end-to-end ST models in both settings.

\noindent\textbf{Comparison with pipeline ST systems.} 
We compare our end-to-end STAST model with text-based MT models and pipeline ST systems. The text-based MT model takes manual transcriptions as input, so its result can be regarded as the upper bound of the speech translation task. For the pipeline system, the ASR model and the NMT model are trained only by data in the ST corpus. The final translation is generated based on the output of the ASR model. As shown in Table \ref{tbl:1}, in the base setting our STAST model can achieve comparable or even better results than pipeline systems under the same data scale. However, STAST only uses a single model to combine ASR and MT tasks, which has much fewer parameters and computational cost than pipeline systems. When more ASR data is available, our method outperforms the best pipeline system by 0.9 BLEU and is slightly worse than text-based MT models. It indicates that with the use of extra ASR corpus, our method can obtain more acoustic information, which has benefits of extracting semantic knowledge and achieving better performances.

\noindent\textbf{Comparison with end-to-end models.}
\label{sec:end-to-end result}
As shown in Table \ref{tbl:1}, our model outperforms all the previous end-to-end models and achieves the new state-of-the-art performance. Specifically, in the base setting our proposed model is much better than the LSMT-based ST model \cite{berard2018end} and outperforms it by $4\sim 5$ BLEU scores. Our model is also better than a widely used toolkit ESPnet by 1.1 BLEU, whose encoder and decoder are both pre-trained. Compared with~\citeauthor{liu2019end}~\shortcite{liu2019end} who utilize an NMT model to teach an ST model on the probability of decoder output, our method achieves 0.8 better BLEU scores. We believe this is because their method only adopts knowledge distillation on the last output layer of model, which is hard to propagate the gradient back to the preceding networks. However, our method adopts cross-modal adaptation on speech and text representation, giving the model more guidance to bridge the gap between two modalities. STAST is also better than TCEN-LSTM proposed by~\citeauthor{wang2020bridging}~\shortcite{wang2020bridging}. Compared with their method which requires to transform normal source sentences to noisy sentences in CTC path format, our method solves length inconsistency problem more effectively. Besides, the speech representation can obtain better semantic information with the cross-modal adaptation. Our method is also superior to the best previous model proposed by~\citeauthor{wang2020curriculum}~\shortcite{wang2020curriculum}. 
In the expanded setting, with extra ASR data, our method improves the translation performance by 0.9 BLEU compared with that in the base setting. It also outperforms other previous end-to-end models by a large margin, which indicates the effectiveness of our method.

\begin{table}[t] \small
\begin{center}
	\begin{tabular}{c|l|c}
		\hline
		Model   &   \multicolumn{1}{|c|}{Method}            &   BLEU  \\
		\hline
		\multirow{3}{*}{MT}  &  LSTM\cite{berard2018end}    & 19.30  \\
		& ESPnet\cite{inaguma2020espnet}   & 18.30  \\
		& Transformer\cite{liu2019end}       & 22.05 \\
		\hline
		\multirow{3}{*}{Pipeline}  & LSTM\cite{berard2018end}    & 14.60  \\
		& ESPnet \cite{inaguma2020espnet}    & 16.96 \\
		& Transformer\cite{liu2019end}         & 17.85 \\
		\hline
		\multirow{9}{*}{\begin{tabular}[c]{@{}lc@{}}E2E \\Base\end{tabular}}
		& LSTM\cite{berard2018end}   & 12.90  \\
		& \quad +pre-train+multitask & 13.40  \\
		& ESPnet \cite{inaguma2020espnet}   & 16.70  \\
		& Transformer \cite{liu2019end}       & 14.30  \\
		& \quad +decoder KD       & 17.02 \\
		& TCEN-LSTM \cite{wang2020bridging}      & 17.05 \\
		& Transformer \cite{wang2020curriculum}       & 15.97 \\
		& \quad +curriculum     & 17.66 \\
		& STAST (ours)    &  \textbf{17.81} \\
		\hline
		\multirow{5}{*}{\begin{tabular}[c]{@{}c@{}} E2E \\ Expanded\end{tabular}} 
		& LSTM\cite{bahar2019on} & 17.00 \\
		& Multilingual\cite{inaguma2020espnet} & 17.60 \\
		& Transformer\cite{wang2020curriculum} & 16.90 \\
		& \quad +curriculum learning & 18.01 \\
		& STAST (ours) & \textbf{18.74} \\
		\hline
	\end{tabular}
\end{center}
\caption{BLEU results on the Augmented LibriSpeech En-Fr corpus.}
\label{tbl:1}
\end{table}

\noindent\textbf{Results on MuST\_C English-German.}
The results on MuST\_C English-German corpus are listed in Table \ref{tbl:2}. 
Compared with the Augmented LibriSpeech corpus, the performance gap between text-based MT models and pipeline systems is much larger. This can be attributed to that the speech in this corpus is recorded from live presentations, which contains more noise, such as laughter, applause, and cheers, besides that the segment of this corpus is also noisy. 
As seen in Table \ref{tbl:2}, STAST model outperforms all of the previous end-to-end models in the base setting. We also find that the STAST model is slightly better than the pipeline system we implement but worse than the best system, which indicates the end-to-end ST model still lacks robustness to the noisy input. It is challenging to handle the speech under noisy circumstances and needs to design more powerful ST models.

\begin{table}[t] \small
\begin{center}
	\begin{tabular}{c|l|c}
		\hline
		Model   &  \multicolumn{1}{|c|}{Method}            &   BLEU  \\
		\hline
		\multirow{2}{*}{MT} 
		& Transformer\cite{gangi2019adapting} & 25.30 \\
		& ESPnet \cite{inaguma2020espnet}   & 30.16  \\
		\hline
		\multirow{2}{*}{Pipeline}   
		& Transformer\cite{gangi2019adapting} & 18.50 \\
		& Transformer \cite{indurthi2020end} &  20.86 \\
		& ESPnet \cite{inaguma2020espnet}    & 23.65 \\
		& Transformer (ours)         & 23.00 \\
		\hline
		\multirow{6}{*}{\begin{tabular}[c]{@{}lc@{}}E2E \\ Base\end{tabular}}        
		& Transformer\cite{gangi2019adapting}   & 17.30  \\
		& Transformer \cite{indurthi2020end}  & 15.60 \\
		& \quad + Meta-Learning & 22.11 \\
		& ESPnet \cite{inaguma2020espnet}  & 22.33  \\
		& \quad + SpecAugment & 22.91 \\
		& STAST (ours)    &  22.55 \\
		& \quad + SpecAugment\tablefootnote{For fair comparison with previous work, we apply SpecAugment \cite{park2019specaugment} method.} & \textbf{23.06} \\
		\hline
	\end{tabular}
\end{center}
\caption{BLEU results on MuST\_C En-De.}
\label{tbl:2}
\end{table}

\subsection{Ablation Studies}
\begin{table}[t] \small
\begin{center}
	\begin{tabular}{l|c|c}
		\hline
		Methods           & LibriSpeech & MuST\_C  \\
		\hline
		STAST        & 18.74    & 22.55 \\
		\quad - Cross-modal Adaptation   & 18.33    & 21.88 \\
		\quad - Multi-task        & 18.17    &  21.25 \\
		\quad - Semantic Encoder & 16.91 & 19.94 \\
		\quad - Shrink Mechanism & 16.55  & 19.10 \\
		\quad - CTC Loss & 16.19  & 18.84 \\
		\hline
	\end{tabular}
\end{center}
\caption{Ablation Studies on the test set of Augmented LibriSpeech En-Fr and MuST\_C En-De.}
\label{tbl:3} 
\end{table}
To better evaluate the contribution of different proposed methods, we perform ablation studies on both corpora. The results in Table \ref{tbl:3} show that all of the proposed methods have positive effects, and the benefits of these methods are accumulative. The performance drops if we remove the cross-modal adaption method. Without multi-task learning, the performance further drops 0.2 and 0.6 BLEU scores. It indicates that both cross-modal adaption and multi-task learning have the ability to transform the latent space of speech representation to text representation closely. With the guidance from the text-based NMT model, the ST model can learn more semantic knowledge and obtain better performance. We find if the shrunk output from the acoustic encoder is directly fed into decoder without the semantic encoder, the translation performance drops significantly. This means decoupling the ST encoder is necessary and the semantic encoder supplements the model with more capability to learn semantic information. If we further remove the shrink mechanism, the performance decreases by 0.4 and 0.8 BLEU scores on two corpora, which proves that blank and repeated tokens hinder the alignment learning between output sequence and input sequence. The shrink mechanism has the ability to bridge the length gap between speech and text representation and is indispensable for cross-modal adaptation. The discussion in detail will be depicted in Section~\ref{sec:shrink_mechanism}. 
Finally, the CTC loss as an auxiliary loss is also beneficial, which is consistent with previous work in the ASR fields that CTC loss has the benefit of accelerating convergence and achieving better performance \cite{kim2017joint}. 


\subsection{Analyses}
\label{sec:shrink_mechanism}
\begin{figure}[t]
\centering
\includegraphics[width=0.9\columnwidth]{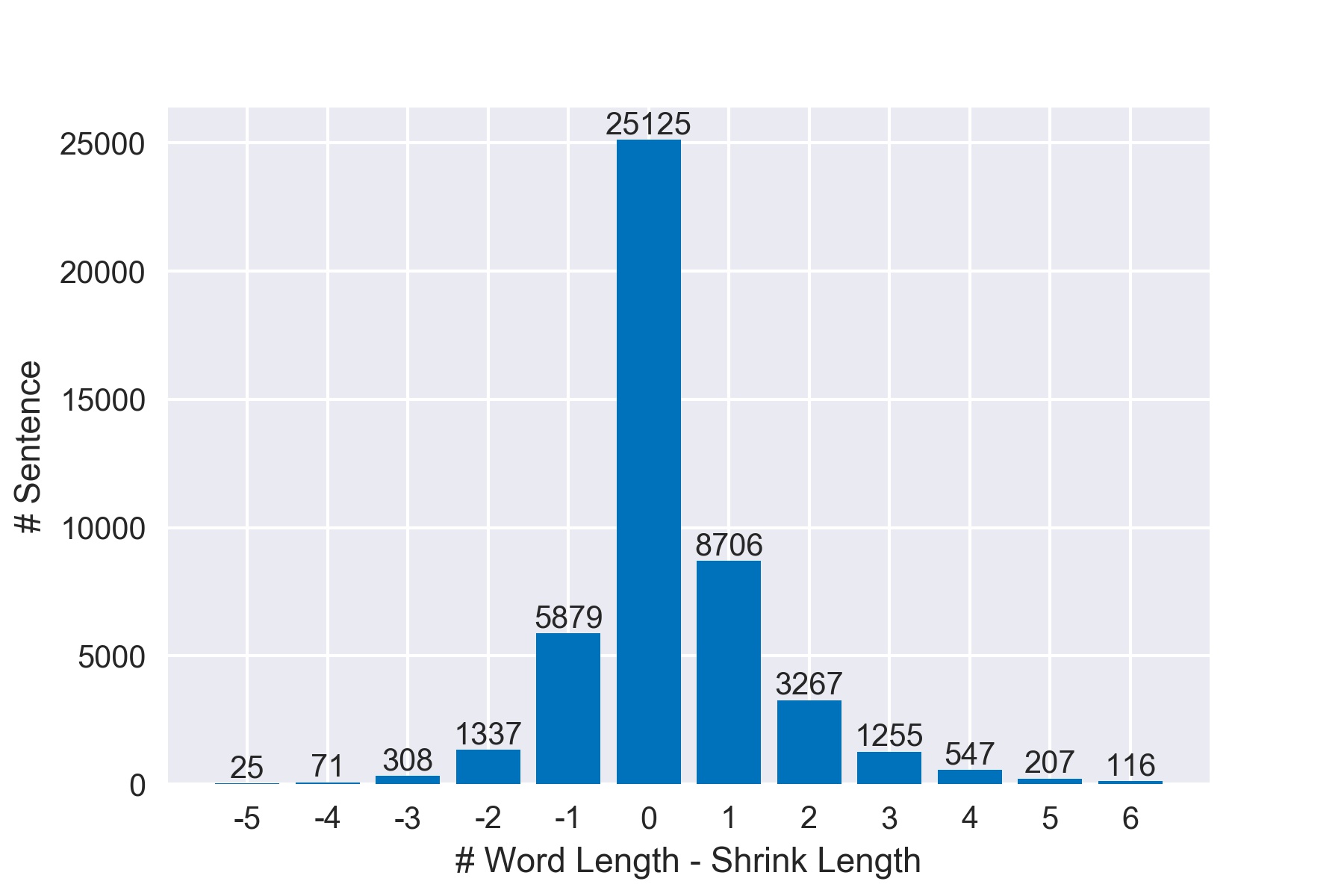}
\caption{The analysis of the length after shrink mechanism, which shows the difference between the length of CTC output after shrink mechanism and that of the source transcription on the training set of Augmented LibriSpeech corpus.}
\label{fig:length}
\end{figure}

\begin{table}[t]
\begin{center}
	\begin{tabular}{l|c|c}
		\hline
		KD Method     & Dev & Test \\ 
		\hline
		Sequence-level  &   19.86     &    18.74      \\
		Word-level &  19.48     &   18.47   \\ 
		\hline
	\end{tabular}
\end{center}
\caption{BLEU results on the development and test set of Augmented LibriSpeech En-Fr with sequence-level adaptation and word-level adaptation.}
\label{tbl:4} 
\end{table}

\textbf{Effect of Shrink Mechanism on the CTC output.}
As shown in Figure \ref{fig:length}, the histogram counts the difference between the length of the CTC output after shrink mechanism and that of the corresponding transcription on the training set of the Augmented LibriSpeech corpus. When the value is below zero, it means the length of shrunk hidden states is longer than that of transcription and vice versa. We find that the length of shrunk hidden states equals to that of the corresponding text transcription for 84.0\% of data, and the difference is less than two for over 93.7\% of data. Therefore, we conclude the shrink mechanism has the ability to solve length inconsistency problem. With more extra ASR data, we believe the accuracy of the shrunk length can be further improved.

\noindent\textbf{Sequence-Level Adaptation v.s. Word-Level Adaptation.}
We compare two different adaptation methods as mentioned in Section \ref{sec:distillation}, i.e. sequence-level adaptation and word-level adaptation. We conduct experiments on the Augmented LibriSpeech corpus and the results are reported in Table \ref{tbl:4}. We find that both adaptation methods have improvements compared with baselines. However, the performance of sequence-level adaptation is slightly better. This is consistent with~\citeauthor{aldarmaki2019context}~\shortcite{aldarmaki2019context}, where they find learning the aggregate mapping can yield a more optimal solution compared to word-level mapping.

\noindent\textbf{Effect of Extra Text Data.}
\begin{figure}[t]
\centering
\includegraphics[width=0.9\columnwidth]{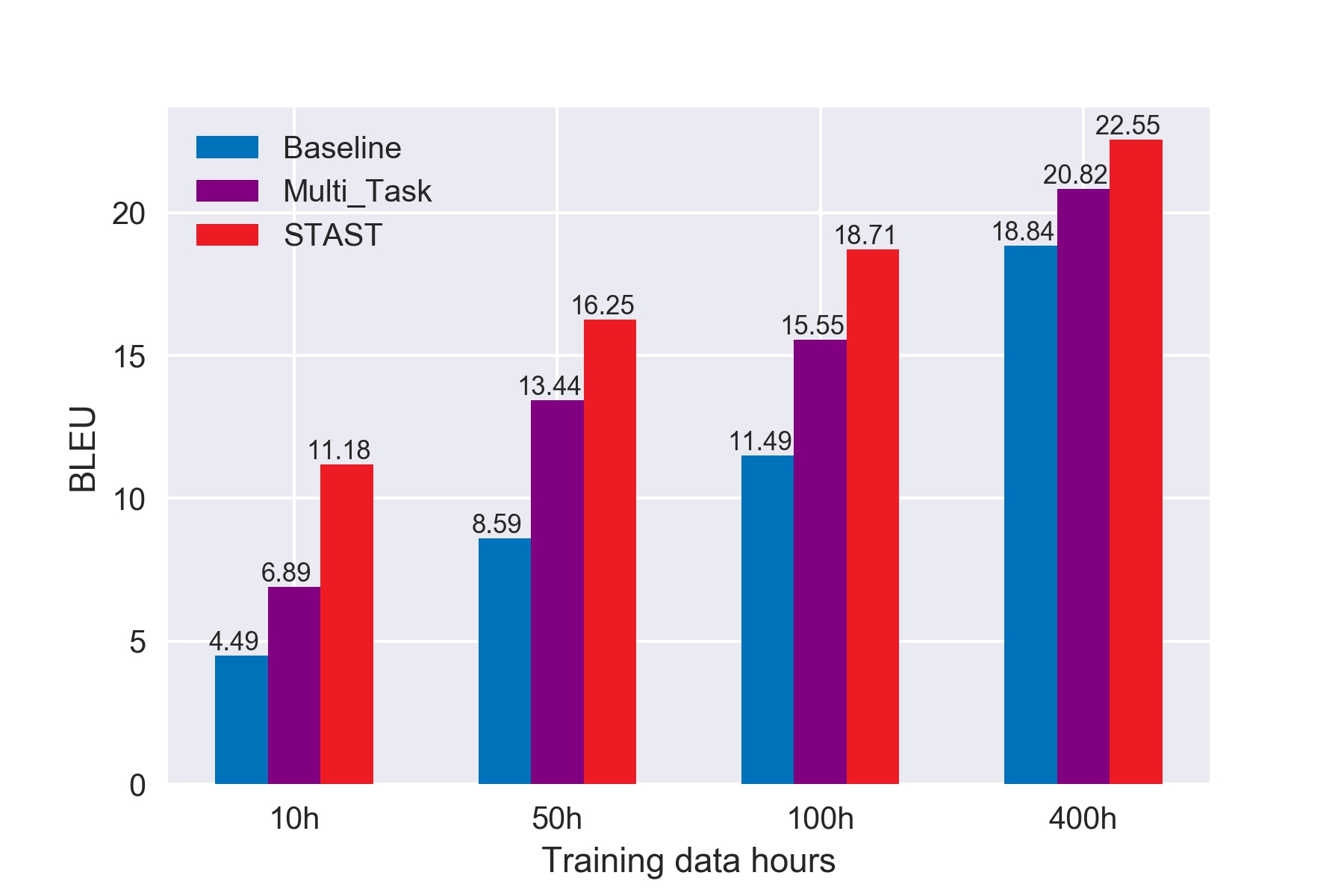}
\caption{The BLEU results of the baseline model, multi-task model and our proposed STAST model on MuST\_C En-De under different scales of training data, including 10 hours, 50 hours, 100 hours, and 400 hours.}
\label{fig:data}
\end{figure}
The module in our proposed STAST model is flexible, which can easily leverage extra data. In Section \ref{sec:end-to-end result}, we have shown that extra ASR data can boost the model performance, here we analyze the effect of extra text data. To simulate lower-resource scenarios, we randomly select 10-hour, 50-hour, and 100-hour of speech training data (speech-transcription-translation triplets) from the MuST\_C En-De dataset, which originally contains 400-hour speech. Then, we train the end-to-end ST baseline model (a simple encoder-decoder model), multi-task baseline model (the decoder of ST and MT is shared with separated encoders), and our STAST model under different scales of training data. The multi-task baseline model and our STAST model can have access to extra text data (transcription-translation pairs in the original corpus). Figure \ref{fig:data} shows the BLEU scores of three models under different scales of training data. We find that with the increase of the data size, all of the three models can obtain improvements. However, the end-to-end model is significantly inferior to the multi-task model and the STAST model. With only 10 hours of training data, our proposed model can achieve comparable performance with the end-to-end model trained on 100-hour data.
The multi-task model is better than the end-to-end model but worse than our proposed model. The reason is that the multi-task model only shares part of parameters for different tasks but leaves the valuable semantic information learned by the NMT encoder unexploited by the ST model. While our method integrates the NMT model into the ST model and transfers speech representation to text representation more closely by the cross-modal adaptation method, which can obtain better performance.

\section{Conclusions}
In this paper, we propose the STAST model to improve end-to-end ST model. Considering the modality differences between speech and text, we propose ST encoder decoupling, length shrink mechanism, the NMT model integration, and cross-modal adaptation methods. Empirical studies have demonstrated that each proposed method has positive effects and the combination of them can achieve the new state-of-the-art result. In the future, we will explore how to effectively transfer more knowledge from the NMT model and the ST model. We also expect that the idea of bridging the representation gap between different modalities can be adopted on other tasks.

\bibliography{acl2020,anthology}
\bibliographystyle{acl_natbib}

\end{document}